\documentclass{article}

\PassOptionsToPackage{numbers, compress}{natbib}

\usepackage[final]{neurips_glfrontiers_2023}

\usepackage[utf8]{inputenc} 
\usepackage[T1]{fontenc}    
\usepackage{hyperref}       
\usepackage{url}            
\usepackage{booktabs}       
\usepackage{amsfonts}       
\usepackage{nicefrac}       
\usepackage{microtype}      
\usepackage{amsmath}
\usepackage{graphicx}
\usepackage{array}
\usepackage[table]{xcolor}

\definecolor{mygray}{RGB}{169, 169, 169}
\definecolor{mypink}{RGB}{245, 171, 186}
\definecolor{myblue}{RGB}{91, 208, 250}
\newcolumntype{C}[1]{>{\centering\arraybackslash}p{#1}}

\definecolor{mypurple}{HTML}{B55690}
\definecolor{myorange}{HTML}{EF7627}
\hypersetup{
  colorlinks=true,          
  linkcolor=myorange,           
  citecolor=mypurple,          
}

\title{Hierarchical Relationships: A New Perspective to Enhance Scene Graph Generation}

\author{%
  Bowen Jiang, Camillo J. Taylor \\
  University of Pennsylvania\\
  Philadelphia, PA, 19104, USA \\
  \texttt{\{bwjiang, cjtaylor\}@seas.upenn.edu}
}

\begin{document}

\maketitle

\begin{abstract}
This paper presents a finding that leveraging the hierarchical structures among labels for relationships and objects can substantially improve the performance of scene graph generation systems.
The focus of this work is to create an informative hierarchical structure that can divide object and relationship categories into disjoint super-categories in a systematic way.
Specifically, we introduce a Bayesian prediction head to jointly predict the super-category of relationships between a pair of object instances, as well as the detailed relationship within that super-category simultaneously, facilitating more informative predictions. The resulting model exhibits the capability to produce a more extensive set of predicates beyond the dataset annotations, and to tackle the prevalent issue of low annotation quality. 
While our paper presents preliminary findings, experiments on the Visual Genome dataset show its strong performance, particularly in predicate classifications and zero-shot settings, that demonstrates the promise of our approach.
\end{abstract}

\section{Introduction}

This work considers the scene graph generation problem~\cite{johnson2015image, lu2016visual, xu2017scene, yang2018graph, zellers2018neural, koner2020relation, suhail2021energy} that deduces the objects in an image and their interconnected relationships. Unlike object detectors~\cite{redmon2016you, ren2015faster, carion2020end} which focus on individual object instances, scene graph models represent the entire image as a graph, where each object instance is a node and the relationship between a pair of nodes is a directed edge.

Existing literature has addressed the nuanced relationships among objects within visual scenes by designing complicated architectures~\cite{dhingra2021bgt, lin2020gps, xu2021joint, cong2022reltr, ren2020scene, chen2019knowledge}, but a gap remains due to the neglect of inherent hierarchical information in relationship categories, resulting in an incomplete understanding of object interactions. We adopt the definitions in Neural Motifs~\cite{zellers2018neural} to divide predominant relationships in scene graphs into \textit{geometric}, \textit{possessive}, and \textit{semantic} super-categories and show how these categories can be explicitly utilized in a network.


In this work, we propose a novel classification scheme inspired by Bayes' rule, jointly predicting relationship super-category probabilities and conditional probabilities of relationships within each super-category. For each directed edge, the top one predicate under each super-category will participate in the ranking, although we still evaluate the recall scores within the top $k$ most confident predicates in each image. Experimental results on the Visual Genome dataset~\cite{krishna2017visual} demonstrate that incorporating hierarchical relationship reasoning can enhance the performance of a baseline model by a large margin, indicating a promising and interesting preliminary finding.

\section{Scene Graph Construction}

\begin{figure}[t]
  \centering
  \includegraphics[width=1\linewidth]{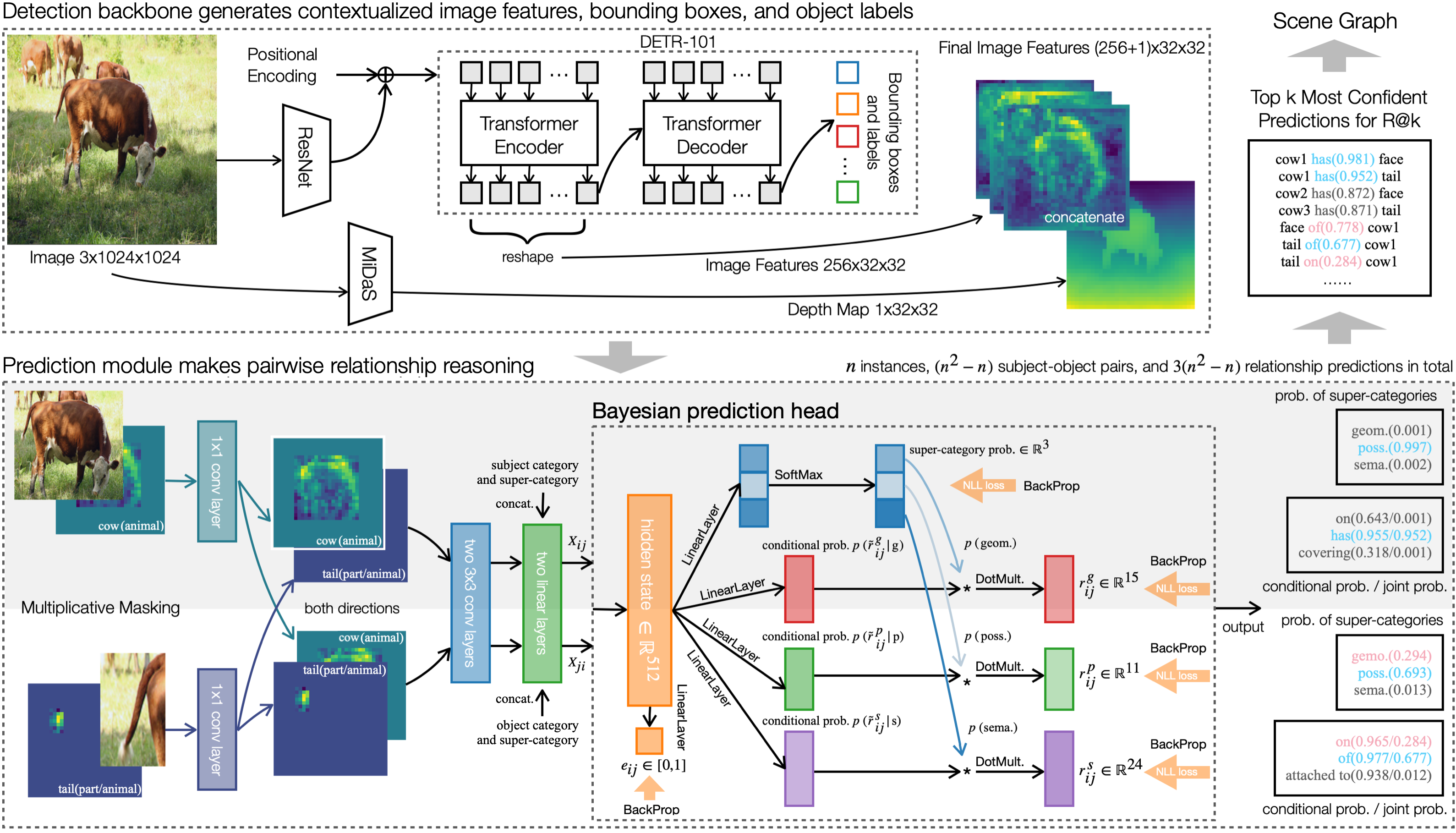}
  \caption{Illustration of the model architecture we construct to showcase the effectiveness of our Bayesian classification head that utilizes the relationship hierarchies. It performs pairwise relationship reasoning for each directed edge in the graph and predicts the probability distributions of both relationship super-categories and detailed categories within each super-category.}
  \vspace{-2mm}
  \label{fig:flowchart}
\end{figure}

\paragraph{Object detection backbone}
Our system adopts the widely-used two-stage design~\cite{xu2017scene, yang2018graph, zellers2018neural, lin2020gps, dhingra2021bgt}. We leverage the Detection Transformer (DETR)~\cite{carion2020end} to predict object bounding boxes and labels. It has a ResNet-101 feature extraction backbone~\cite{he2016deep} and a transformer encoder~\cite{carion2020end, vaswani2017attention} that contextualizes the feature space with global information. Its output still preserves its spatial dimensions and serves as the image features $\boldsymbol{I} \in \mathbb{R}^{h \times s \times t}$, where $h$ is the hidden dimensions, and $s$ and $t$ denote its spatial dimensions. We also employ the MiDaS~\cite{ranftl2020towards} single-image depth estimation network to provide depth maps $\boldsymbol{D}$ for input images. The final image features $\boldsymbol{I}^{\prime} = \operatorname{concat}\{\boldsymbol{I}, \boldsymbol{D}\} \in \mathbb{R}^{(h + 1)\times s \times t}$ serve as the inputs to the subsequent networks for relationship reasoning.

\paragraph{Direction-aware masking}
After extracting $\boldsymbol{I}^{\prime}$, the model considers each pair of object instances. We construct a combined feature tensor by masking $\boldsymbol{I}^{\prime}$ with the bounding boxes $\boldsymbol{M}_{i}, \boldsymbol{M}_{j} \in\mathbb{R}^{s\times t}$ of the subject and object, resulting in two feature tensors $\boldsymbol{I}_{i}^{\prime}, \boldsymbol{I}_{j}^{\prime} \in\mathbb{R}^{(h+1)\times s \times t}$. The order of these two tensors matters.
For example, choosing $\textit{<bike, has, wheel>}$ or $\textit{<wheel, of, bike>}$ depends on which instance is considered as the subject and which one as the object. Therefore, we avoid using a union mask of the subject-object pair but instead, perform two separate passes through the model, depicted in Figure~\ref{fig:flowchart}. One pass considers $\boldsymbol{I}_{i}^{\prime}$ as the subject and concatenates them as $\boldsymbol{I}_{ij}^{\prime}$, while the other pass swaps their roles to yield $\boldsymbol{I}_{ji}^{\prime}$. Each one is fed to subsequent convolutional layers, flattened, concatenated with four one-hot vectors encoding the categories and super-categories associated with the subject and the object, and reduced to a 512-dimensional hidden vector denoted as $\boldsymbol{X}_{ij}$. 

\paragraph{Bayesian prediction head}
Inspired by Baye's rule, the head predicts from $\boldsymbol{X}_{ij}$ a scalar connectivity score $0 \le e_{ij} \le 1$, the probability of three relationship super-categories $\boldsymbol{r}_{ij}^{\text{sup}} = [\mathbf{p}(\text{geo}), \ \mathbf{p}(\text{pos}), \ \mathbf{p}(\text{sem})]\in\mathbb{R}^{3}$, and the conditional probability distributions $\{ \boldsymbol{r}_{ij}^{\text{sub}} = \mathbf{p}(\boldsymbol{\boldsymbol{\tilde{r}}}_{ij}^{\text{sub}} | \text{sub}) \mid \text{sub} \in [\text{geo}, \text{pos}, \text{sem}] \}$ under each super-categories, respectively, each of which is computed by multiplying its conditional probability vector with the associated super-category probability.
\begin{align}
	e_{ij} &= \operatorname{Sigmoid}\left\{\boldsymbol{X}_{ij}^{\top}\boldsymbol{W}^{\text{conn}}\right\} \label{eqn:conn} \\
	\boldsymbol{r}_{ij}^{\text{sup}} &= [\mathbf{p}(\text{geo}), \ \mathbf{p}(\text{pos}), \ \mathbf{p}(\text{sem})] = \operatorname{SoftMax}\left\{\boldsymbol{X}_{ij}^{\top} \boldsymbol{W}^{\text{sup}}\right\} \\
	\boldsymbol{r}_{ij}^{\text{sub}} &= \operatorname{SoftMax} \left\{\boldsymbol{X}_{ij}^{\top} \boldsymbol{W}^{\text{sub}} * \mathbf{p}(\text{sub}) \right\} \text{, for sub} \in [\text{geo}, \text{pos}, \text{sem}], \label{eqn:post}
\end{align}
where $*$ is the scalar product and all $\boldsymbol{W}$s are the learnable parameter tensors of linear layers. 

To train the Bayesian classification head, we apply one cross-entropy loss to the super-categories and another cross-entropy loss to the detailed relationships under the ground-truth super-category. Furthermore, we use a supervised contrastive loss~\cite{khosla2020supervised} shown in Equation~\ref{eqn:contrast} to minimize the distances between hidden states corresponding to the same relation class (set $P(ij)$), while maximizing the distances between those from different relationship classes (set $N(ij)$).
\begin{equation}
\mathcal{L}_{\text{contrast}} \sim \sum_{p \in P(ij)}\hspace{-1mm} \log \frac{\exp \left(\boldsymbol{X}_{ij}^{\top} \boldsymbol{X}_{p} / \tau \right)}{\sum_{n \in N(ij)} \exp \left(\boldsymbol{X}_{ij}^{\top} \boldsymbol{X}_{n} / \tau\right)}, \text{ where } \tau \text{ is the temperature.}
\label{eqn:contrast}
\end{equation}
\vspace{-1.5mm}

The model yields three predicates on each edge, one from each disjoint super-category, which maintains the exclusivity among predicates within the same super-category. While keeping the same evaluation metrics that focus on recall scores within the top $k$ most confident predicates in an image, all three predicates from each edge will participate in the ranking. Because there will be three times more candidates, we are not trivially relieving the graph constraints~\cite{newell2017pixels, zellers2018neural} to make the task simple.

We find it common for one or two predicates from the same edge to appear within the top $k$ of the ranking, providing potentially matched solutions at disjoint super-categories to enhance the performance, while pushing more valueless predicates out of the rank. This design leverages the super-category probabilities to guide the network's attention toward the appropriate conditional output heads, enhancing the interpretability and performance of the system. Furthermore, the use of separate but smaller linear layers does not exacerbate the number of parameters or scalability issues.

\vspace{-2mm}
\section{Experiments}

\paragraph{Dataset and training techniques}
Our experiments are conducted on the Visual Genome dataset~\cite{krishna2017visual}, following the same pre-processing procedures outlined in~\cite{xu2017scene}. We filter out the top 150 object categories and 50 relationship predicates, resulting in nearly $75.7k$ training images and $32.4k$ testing images. We adopt the DETR pretrained by~\cite{li2022sgtr} and freeze its parameters throughout all our experiments. The subsequent model is trained using SGD with a learning rate of 1e-5, a step scheduler to reduce the learning rate by 10 at the third epoch, and a batch size of 16 for 3 epochs on four NVIDIA V100 GPUs. The average inference time is 0.16 seconds per image.
\vspace{-1mm}

\paragraph{Evaluation metrics}
We assess performance using R@$k$ and mR@$k$ metrics~\cite{lu2016visual, tang2019learning}. R@$k$ measures the recall within the top $k$ most confident predicates per image, while mR@$k$ computes the average across all relationship classes. We conduct three tasks: (1) Predicate classification (PredCLS) predicts relationships with known bounding boxes and labels. (2) Scene graph classification (SGCLS) only assumes known bounding boxes. (3) Scene graph detection (SGDET) has no prior knowledge.
\vspace{-1mm}

\paragraph{Results}
We compare our model with state-of-the-art methods and include preliminary ablation studies in Table~\ref{tab:result} and \ref{tab:ablation}. The most important comparison is between ``ours" and ``ours [a]" in Table~\ref{tab:result}, where we compare a model with a flat classification head and a model with the Bayesian classification head. The flat model without the hierarchical structure achieves inferior scores, as expected. However, despite the simplicity of our architectural design, adding the hierarchical relationships enhances the scores by a large margin particularly on the PredCLS task and allows our model to achieve a competitive performance. The approach also exhibits strong zero-shot performance in Table~\ref{tab:zero}.

\begin{figure}[t]
  \centering
  \includegraphics[width=1\linewidth]{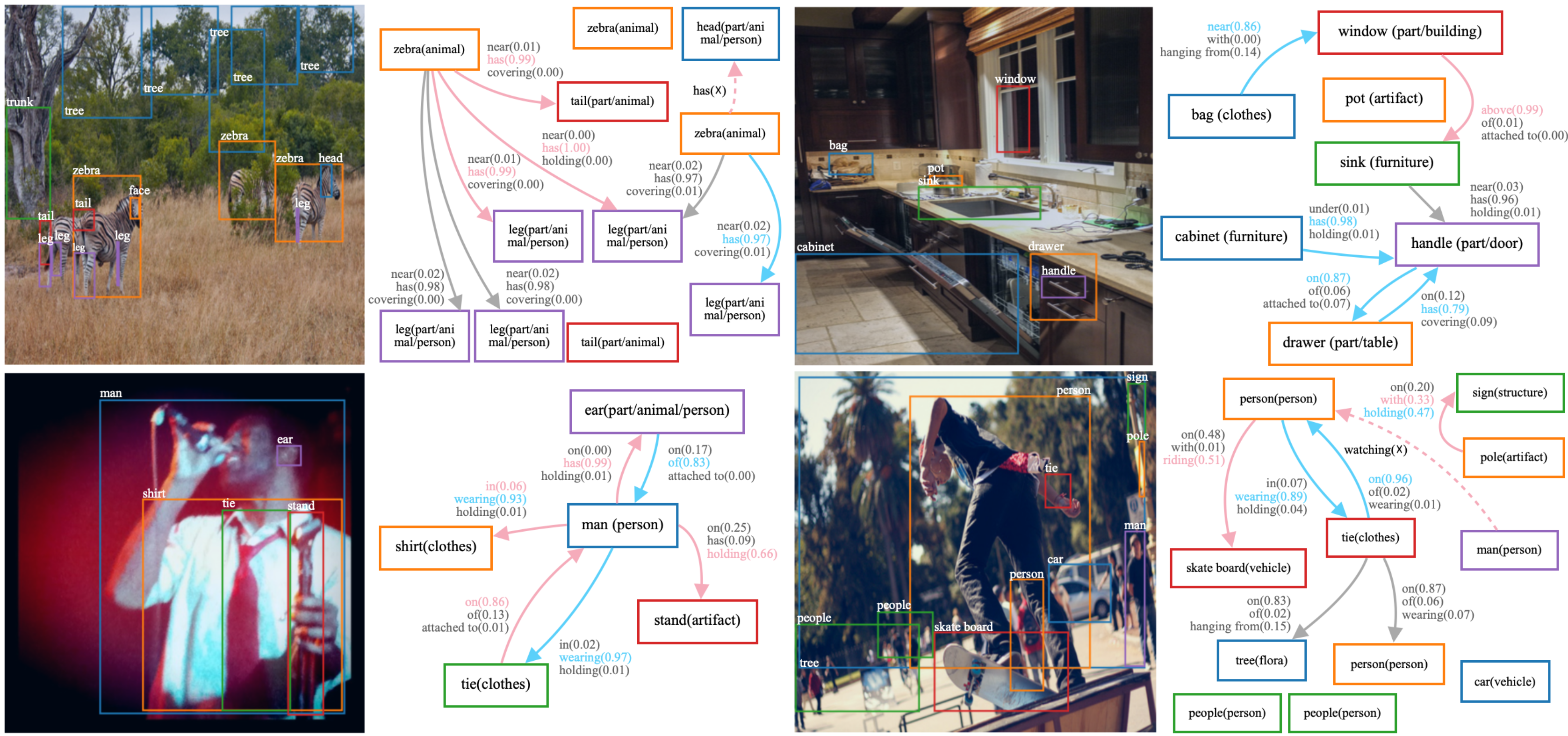}
  \caption{Examples of generated scene graphs. We only show the most confident edges with their top one predicates under all three super-categories and the super-category probabilities. We sketch (1) solid \textcolor{mypink}{pink} arrow: true positives. (2) dotted \textcolor{mypink}{pink} arrow: false negatives. (3) solid \textcolor{myblue}{blue} arrow: reasonably true positives not yet annotated in the dataset. (4) solid \textcolor{mygray}{gray} arrow: false positives.}
  \label{fig:examples}
\end{figure}

Figure~\ref{fig:examples} illustrates some predicted scene graphs. Each edge is annotated with the top predicate from three distinct super-categories. It is intriguing to discover that there are numerous reasonable predicates aligned with human intuitions but not annotated in the dataset, marked in \textcolor{myblue}{blue}. There are also scenarios where the top predicate under the second most likely super-category is picked by the dataset as the ground truth. Conventional algorithms would classify this edge as a false negative, but our approach acknowledges it as long as it still appears within the top $k$ of the ranking. We strongly believe that creating an extensive set of predicates is beneficial for practical scene understanding.

\begin{table*}
\vspace{-3mm}
  \centering
  \footnotesize
  \caption{Test results and ablation studies. [a] means hierarchical relationship.}
  \label{your-table}
  \begin{tabular}{
        *{1}{C{2cm}}
        *{3}{C{0.83cm}}
        *{3}{C{0.83cm}}
        *{3}{C{0.83cm}}
      }
    \\ \toprule
    \multicolumn{1}{c}{} &
    \multicolumn{3}{c}{PredCLS} &
    \multicolumn{3}{c}{SGCLS} &
    \multicolumn{3}{c}{SGDET} \\
    \cmidrule(r){2-4}
    \cmidrule(r){5-7}
    \cmidrule(r){8-10}
    Methods & R@20 & R@50 & R@100 & R@20 & R@50 & R@100 & R@20 & R@50 & R@100 \\
    \midrule
    NeuralMotifs~\cite{zellers2018neural} & 58.5 & 65.2 & 67.1 & 32.9 & 35.8 & 36.5 & 21.4 & 27.2 & 30.3 \\
    HC-Net~\cite{ren2020scene} & 59.6 & 66.4 & 68.8 & 34.2 & 36.6 & 37.3 & 22.6 & 28.0 & 31.2 \\
    GPS-Net~\cite{lin2020gps} & 60.7 & 66.9 & 68.8 & 36.1 & 39.2 & 40.1 & 22.6 & 28.4 & 31.7 \\
    BGT-Net~\cite{dhingra2021bgt} & 60.9 & 67.3 & 68.9 & 38.0 & 40.9 & 43.2 & 23.1 & 28.6 & 32.2 \\
    RelTR~\cite{cong2022reltr} & 63.1 & 64.2 & - & 29.0 & 36.6 & - & 21.2 & 27.5 & - \\
    \rowcolor{gray!10} 
    \hline
    ours & 61.1 & 73.6 & 78.1 & 30.6 & 36.0 & 37.6 & 22.8 & 29.3 & 32.1  \\
    \rowcolor{gray!10} 
    ours w/o \verb|[a]| & 56.6 & 66.6 & 69.1 & 29.5 & 33.8 & 34.8 & 20.2 & 26.3 & 28.1 \\
    \toprule
    & mR@20 & mR@50 & mR@100 & mR@20 & mR@50 & mR@100 & mR@20 & mR@50 & mR@100 \\
    NeuralMotifs~\cite{zellers2018neural} & 11.7 & 14.8 & 16.1 & 6.7 & 8.3 & 8.8 & 4.9 & 6.8 & 7.9 \\
    Motif+EB~\cite{suhail2021energy} & 14.2 & 18.0 & 19.5 & 8.2 & 10.2 & 10.9 & 5.7 & 7.7 & 9.3 \\
    GPS-Net~\cite{lin2020gps} & 17.4 & 21.3 & 22.8 & 10.0 & 11.8 & 12.6 & 6.9 & 8.7 & 9.8 \\
    BGT-Net~\cite{dhingra2021bgt} & 16.8 & 20.6 & 23.0 & 10.4 & 12.8 & 13.6 & 5.7 & 7.8 & 9.3 \\
    RelTR~\cite{cong2022reltr} & 20.0 & 21.2 & - & 7.7 & 11.4 & - & 6.8 & 10.8 & - \\
    \rowcolor{gray!10} 
    \hline
    ours & 14.4 & 20.6 & 23.7 & 7.7 & 10.4 & 11.9 & 4.1 & 6.8 & 8.7 \\
    \rowcolor{gray!10} 
    ours w/o \verb|[a]| & 9.5 & 14.5 & 14.9 & 5.8 & 7.2 & 7.8 & 3.2 & 4.6 & 5.4 \\
    \bottomrule
  \end{tabular}
  \label{tab:result}
\end{table*}

\begin{table}[h!]
  \vspace{-1.5mm}
  \centering
  \label{tab:result2}
  \begin{minipage}{0.5\textwidth}
    \centering
    \footnotesize
    \caption{More ablation studies (PredCLS). [b] ours w/o the depth maps. [c] ours w/o the supervised contrastive loss.}
    \label{tab:ablation}
    \begin{tabular}{
          *{1}{c}
          *{4}{c}
        }
      \toprule
      Methods & R@20 & R@50 & mR@20 & mR@50 \\
      \midrule
      \verb|[b]| & 62.5 & 74.2 & 15.5 & 21.6 \\
      \verb|[c]| & 62.1 & 74.7 & 14.8 & 20.4 \\
      \bottomrule
    \end{tabular}
  \end{minipage}
  \begin{minipage}{0.49\textwidth}
    \centering
    \footnotesize
    \caption{Zero-shot recall~\cite{tang2020unbiased} (PredCLS).}
    \label{tab:zero}
    \begin{tabular}{
          *{1}{c}
          *{2}{c}
        }
      \toprule
      Methods & zsR@20 & zsR@50 \\
      \midrule
      NeuralMotifs~\cite{zellers2018neural} & 1.4 & 3.6 \\
      Motif+EB~\cite{suhail2021energy} & 2.1 & 4.9 \\
      VC-TDE+EB~\cite{suhail2021energy} & 9.6 & 15.1 \\
      \midrule
      ours & 10.9 & 20.4 \\
      ours w/o \verb|[a]| & 9.4 & 14.7 \\
      \bottomrule
    \end{tabular}
  \end{minipage}
  \vspace{-3.5mm}
\end{table}

\section{Discussion and Future Work}
We present a straightforward yet powerful scene graph generation algorithm that effectively exploits relationship hierarchies. 
The resulting system adopts a Bayesian prediction head, enabling simultaneous prediction of the super-categories and specific relationships within each super-category. 
This preliminary study suggests that factorizing the final probability distribution over the relationship categories could enhance the scene graph generation performance, and produce a diverse set of predicates beyond dataset annotations. It also shows strong zero-shot performance.
In the near future, we are going to perform more comprehensive ablation studies and experiments on different datasets, and most importantly, extend our hierarchical classification scheme, as a portable module, to other existing state-of-the-art scene graph generation algorithms and enhance their results.

\section{Acknowledge}
This work was supported by the National Science Foundation (NSF) grant CCF-2112665 (TILOS).

{\small
    \bibliographystyle{plain}
    \bibliography{refs}

\begin{thebibliography}{10}

\bibitem{carion2020end}
Nicolas Carion, Francisco Massa, Gabriel Synnaeve, Nicolas Usunier, Alexander Kirillov, and Sergey Zagoruyko.
\newblock End-to-end object detection with transformers.
\newblock In {\em European conference on computer vision}, pages 213--229. Springer, 2020.

\bibitem{chen2019knowledge}
Tianshui Chen, Weihao Yu, Riquan Chen, and Liang Lin.
\newblock Knowledge-embedded routing network for scene graph generation.
\newblock In {\em Proceedings of the IEEE/CVF Conference on Computer Vision and Pattern Recognition}, pages 6163--6171, 2019.

\bibitem{cong2022reltr}
Yuren Cong, Michael~Ying Yang, and Bodo Rosenhahn.
\newblock Reltr: Relation transformer for scene graph generation.
\newblock {\em arXiv preprint arXiv:2201.11460}, 2022.

\bibitem{dhingra2021bgt}
Naina Dhingra, Florian Ritter, and Andreas Kunz.
\newblock Bgt-net: Bidirectional gru transformer network for scene graph generation.
\newblock In {\em Proceedings of the IEEE/CVF Conference on Computer Vision and Pattern Recognition}, pages 2150--2159, 2021.

\bibitem{he2016deep}
Kaiming He, Xiangyu Zhang, Shaoqing Ren, and Jian Sun.
\newblock Deep residual learning for image recognition.
\newblock In {\em Proceedings of the IEEE conference on computer vision and pattern recognition}, pages 770--778, 2016.

\bibitem{johnson2015image}
Justin Johnson, Ranjay Krishna, Michael Stark, Li-Jia Li, David Shamma, Michael Bernstein, and Li~Fei-Fei.
\newblock Image retrieval using scene graphs.
\newblock In {\em Proceedings of the IEEE conference on computer vision and pattern recognition}, pages 3668--3678, 2015.

\bibitem{khosla2020supervised}
Prannay Khosla, Piotr Teterwak, Chen Wang, Aaron Sarna, Yonglong Tian, Phillip Isola, Aaron Maschinot, Ce~Liu, and Dilip Krishnan.
\newblock Supervised contrastive learning.
\newblock {\em Advances in neural information processing systems}, 33:18661--18673, 2020.

\bibitem{koner2020relation}
Rajat Koner, Suprosanna Shit, and Volker Tresp.
\newblock Relation transformer network.
\newblock {\em arXiv preprint arXiv:2004.06193}, 2020.

\bibitem{krishna2017visual}
Ranjay Krishna, Yuke Zhu, Oliver Groth, Justin Johnson, Kenji Hata, Joshua Kravitz, Stephanie Chen, Yannis Kalantidis, Li-Jia Li, David~A Shamma, et~al.
\newblock Visual genome: Connecting language and vision using crowdsourced dense image annotations.
\newblock {\em International journal of computer vision}, 123(1):32--73, 2017.

\bibitem{li2022sgtr}
Rongjie Li, Songyang Zhang, and Xuming He.
\newblock Sgtr: End-to-end scene graph generation with transformer.
\newblock In {\em Proceedings of the IEEE/CVF Conference on Computer Vision and Pattern Recognition}, pages 19486--19496, 2022.

\bibitem{lin2020gps}
Xin Lin, Changxing Ding, Jinquan Zeng, and Dacheng Tao.
\newblock Gps-net: Graph property sensing network for scene graph generation.
\newblock In {\em Proceedings of the IEEE/CVF Conference on Computer Vision and Pattern Recognition}, pages 3746--3753, 2020.

\bibitem{lu2016visual}
Cewu Lu, Ranjay Krishna, Michael Bernstein, and Li~Fei-Fei.
\newblock Visual relationship detection with language priors.
\newblock In {\em European conference on computer vision}, pages 852--869. Springer, 2016.

\bibitem{newell2017pixels}
Alejandro Newell and Jia Deng.
\newblock Pixels to graphs by associative embedding.
\newblock {\em Advances in neural information processing systems}, 30, 2017.

\bibitem{ranftl2020towards}
Ren{\'e} Ranftl, Katrin Lasinger, David Hafner, Konrad Schindler, and Vladlen Koltun.
\newblock Towards robust monocular depth estimation: Mixing datasets for zero-shot cross-dataset transfer.
\newblock {\em IEEE transactions on pattern analysis and machine intelligence}, 2020.

\bibitem{redmon2016you}
Joseph Redmon, Santosh Divvala, Ross Girshick, and Ali Farhadi.
\newblock You only look once: Unified, real-time object detection.
\newblock In {\em Proceedings of the IEEE conference on computer vision and pattern recognition}, pages 779--788, 2016.

\bibitem{ren2020scene}
Guanghui Ren, Lejian Ren, Yue Liao, Si~Liu, Bo~Li, Jizhong Han, and Shuicheng Yan.
\newblock Scene graph generation with hierarchical context.
\newblock {\em IEEE Transactions on Neural Networks and Learning Systems}, 32(2):909--915, 2020.

\bibitem{ren2015faster}
Shaoqing Ren, Kaiming He, Ross Girshick, and Jian Sun.
\newblock Faster r-cnn: Towards real-time object detection with region proposal networks.
\newblock {\em Advances in neural information processing systems}, 28, 2015.

\bibitem{suhail2021energy}
Mohammed Suhail, Abhay Mittal, Behjat Siddiquie, Chris Broaddus, Jayan Eledath, Gerard Medioni, and Leonid Sigal.
\newblock Energy-based learning for scene graph generation.
\newblock In {\em Proceedings of the IEEE/CVF conference on computer vision and pattern recognition}, pages 13936--13945, 2021.

\bibitem{tang2020unbiased}
Kaihua Tang, Yulei Niu, Jianqiang Huang, Jiaxin Shi, and Hanwang Zhang.
\newblock Unbiased scene graph generation from biased training.
\newblock In {\em Proceedings of the IEEE/CVF conference on computer vision and pattern recognition}, pages 3716--3725, 2020.

\bibitem{tang2019learning}
Kaihua Tang, Hanwang Zhang, Baoyuan Wu, Wenhan Luo, and Wei Liu.
\newblock Learning to compose dynamic tree structures for visual contexts.
\newblock In {\em Proceedings of the IEEE/CVF conference on computer vision and pattern recognition}, pages 6619--6628, 2019.

\bibitem{vaswani2017attention}
Ashish Vaswani, Noam Shazeer, Niki Parmar, Jakob Uszkoreit, Llion Jones, Aidan~N Gomez, {\L}ukasz Kaiser, and Illia Polosukhin.
\newblock Attention is all you need.
\newblock {\em Advances in neural information processing systems}, 30, 2017.

\bibitem{xu2017scene}
Danfei Xu, Yuke Zhu, Christopher~B Choy, and Li~Fei-Fei.
\newblock Scene graph generation by iterative message passing.
\newblock In {\em Proceedings of the IEEE conference on computer vision and pattern recognition}, pages 5410--5419, 2017.

\bibitem{xu2021joint}
Minghao Xu, Meng Qu, Bingbing Ni, and Jian Tang.
\newblock Joint modeling of visual objects and relations for scene graph generation.
\newblock {\em Advances in Neural Information Processing Systems}, 34:7689--7702, 2021.

\bibitem{yang2018graph}
Jianwei Yang, Jiasen Lu, Stefan Lee, Dhruv Batra, and Devi Parikh.
\newblock Graph r-cnn for scene graph generation.
\newblock In {\em Proceedings of the European conference on computer vision (ECCV)}, pages 670--685, 2018.

\bibitem{zellers2018neural}
Rowan Zellers, Mark Yatskar, Sam Thomson, and Yejin Choi.
\newblock Neural motifs: Scene graph parsing with global context.
\newblock In {\em Proceedings of the IEEE conference on computer vision and pattern recognition}, pages 5831--5840, 2018.

\end{thebibliography}
}

\end{document}